\documentclass[conference]{IEEEtran}

\usepackage{times}

\usepackage[numbers]{natbib}
\usepackage{multicol}
\usepackage[bookmarks=true]{hyperref}

\usepackage[utf8]{inputenc} 
\usepackage[T1]{fontenc}    
\usepackage{url}            
\usepackage{booktabs}       
\usepackage{amsfonts}       
\usepackage{nicefrac}       
\usepackage{microtype}      
\usepackage{xcolor}         

\usepackage{url}
\usepackage{graphicx}
\usepackage{caption} 
\usepackage{algorithm}
\usepackage[noend]{algpseudocode}
\algrenewcommand{\algorithmiccomment}[1]{\hfill// #1}

\usepackage{amsmath,amssymb}
\usepackage{multirow}
\usepackage{array,multirow,graphicx}
\usepackage{float}
\usepackage{wrapfig}
\usepackage{todonotes}
\usepackage{subcaption}
\usepackage{amsmath}

\usepackage{xparse} 
\usepackage{amsmath} 
\usepackage{amssymb}
\usepackage{amsfonts} 
\usepackage{dsfont}
\usepackage{graphicx}
\usepackage{url}
\usepackage{booktabs}
\usepackage[para]{threeparttable}
\usepackage{multirow}
\usepackage{amsthm}
\usepackage{xcolor}
\usepackage{framed}
\usepackage{caption}
\usepackage{subcaption}

\graphicspath{{figures/}}

\NewDocumentCommand\bbm{}{ \begin{bmatrix} }
\NewDocumentCommand\ebm{}{ \end{bmatrix} }
\NewDocumentCommand\Vector{m}{ \boldsymbol{\mathbf{#1}} }
\NewDocumentCommand\Matrix{m}{ \boldsymbol{\mathbf{#1}} }

\NewDocumentCommand\Real{}{ \mathbb{R} }

\NewDocumentCommand\NormalDistribution{mm}{ \mathcal{N}\left(#1,#2\right) }

\NewDocumentCommand\Identity{}{ \Matrix{I} }

\NewDocumentCommand\Rotation{}{ \Matrix{R} }

\NewDocumentCommand\T{}{\mathsf{T}}

\NewDocumentCommand\diag{m}{\text{diag}\left(#1\right)}
\NewDocumentCommand\BinghamDistribution{}{\textbf{Bingham}}

\newcommand{\bs}{\mathbf{s}}
\newcommand{\ba}{\mathbf{a}}

\newcommand{\st}{\bs_t}
\newcommand{\at}{\ba_t}

\newcommand{\states}{\mathcal{S}}
\newcommand{\actions}{\mathcal{A}}

\newcommand{\qattn}{Q}

\newcommand{\qattnp}{\theta}

\newcommand{\obs}{\mathbf{o}}
\newcommand{\rgb}{\mathbf{b}}

\newcommand{\pcd}{\mathbf{p}}
\newcommand{\proprio}{\mathbf{z}}

\DeclareMathOperator*{\E}{\mathbb{E}}
\DeclareMathOperator*{\argmaxtwod}{argmax\textit{2D}}

\begin{document}

\title{Bingham Policy Parameterization for 3D Rotations in Reinforcement Learning}

\author{
\authorblockN{Stephen James and Pieter Abbeel}
\authorblockA{UC Berkeley\\
\{stepjam,pabbeel\}@berkeley.edu}}


\maketitle

\begin{abstract}
We propose a new policy parameterization for representing 3D rotations during reinforcement learning. Today in the continuous control reinforcement learning literature, many stochastic policy parameterizations are Gaussian. We argue that universally applying a Gaussian policy parameterization is not always desirable for all environments. One such case in particular where this is true are tasks that involve predicting a 3D rotation output, either in isolation, or coupled with translation as part of a full 6D pose output. Our proposed Bingham Policy Parameterization (BPP) models the Bingham distribution and allows for better rotation (quaternion) prediction over a Gaussian policy parameterization in a range of reinforcement learning tasks. We evaluate BPP on the rotation Wahba problem task, as well as a set of vision-based next-best pose robot manipulation tasks from RLBench. We hope that this paper encourages more research into developing other policy parameterization that are more suited for particular environments, rather than always assuming Gaussian. Code available at: \url{https://sites.google.com/view/rl-bpp}.
\end{abstract}

\IEEEpeerreviewmaketitle


\section{Introduction}

Deep reinforcement learning (RL) is now actively used in many areas, including playing games~\cite{mnih2015human, silver2016mastering}, robot manipulation~\cite{matas2018sim, james2021coarse}, and legged robotics~\cite{hwangbo2019learning, rudin2021cat}. The leading (general-purpose) algorithms within the continuous control RL community are either deterministic, such as DDPG~\cite{lillicrap2015continuous} and TD3~\cite{fujimoto2018addressing}, or stochastic, such as SAC~\cite{haarnoja2018soft} and PPO~\cite{schulman2017proximal}. However, many stochastic RL algorithms parameterize their policies to output multivariate Gaussian distribution parameters; that is, they assume that the action space can be represented by sampling from a multivariate Gaussian, though in practice, these also assume a diagonal covariance. Although assuming Gaussian has worked tremendously in a larger number of domains, we argue that easy performance gains are being missed by simply not using more suitable policy parameterizations. One such example is an environment that requires a 3D rotation action space. Quaternions are often used as the output rotation representation when using deep networks~\cite{wang2019densefusion, wada2020morefusion, james2021attention}, however due to the complex bi-modal and symmetric property, sampling a quaternion from a Gaussian does not seem appropriate. 

\begin{figure}
     \centering
     \includegraphics[width=\linewidth]{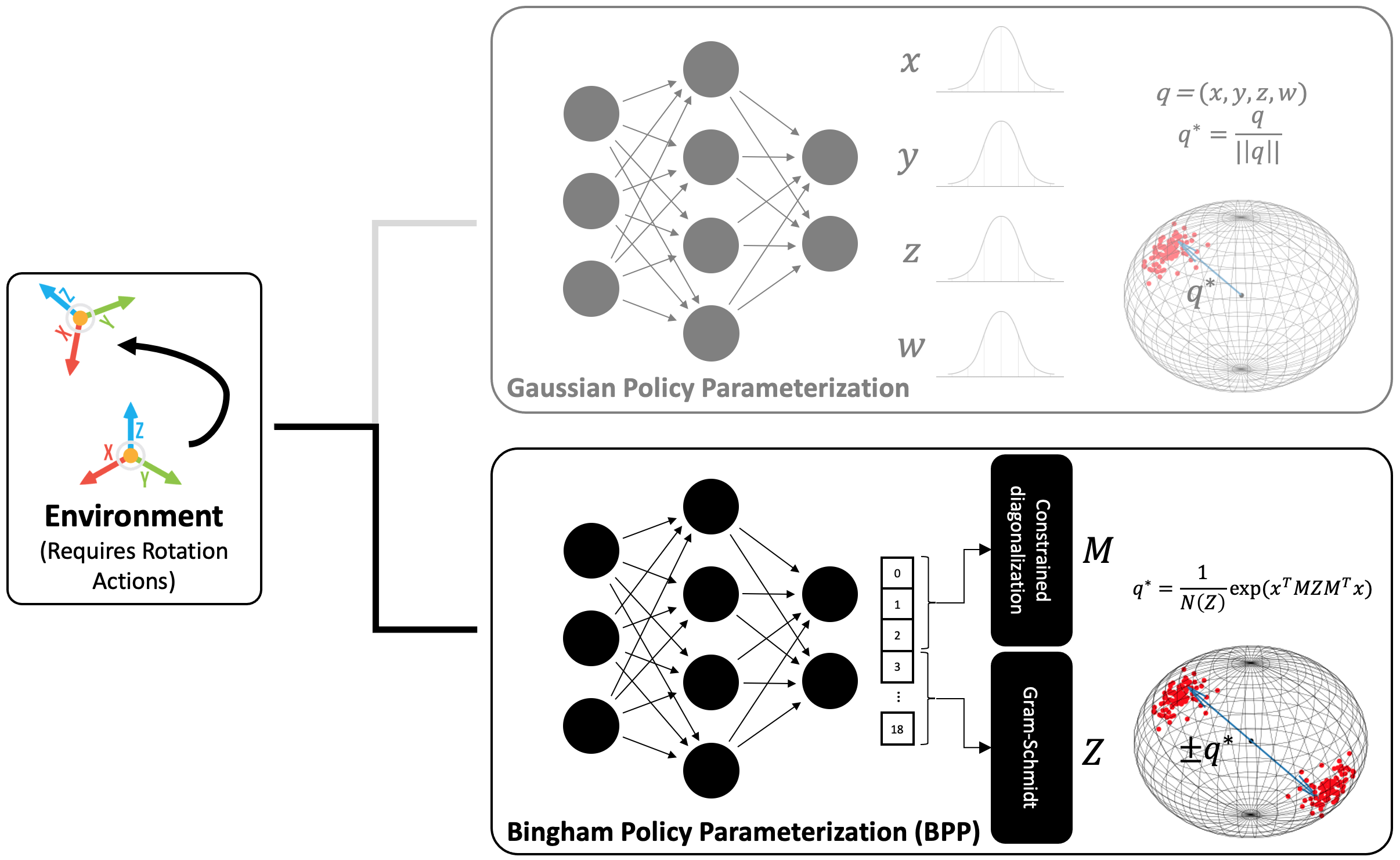}
     \caption{Not all reinforcement learning problems are suited for a Gaussian policy parameterization. In the case of 3D rotation action spaces, swapping to a Bingham policy parameterization (BPP) can yield superior performance.}
     \label{fig:front_summary}
\end{figure}

In this paper, we advocate for re-examining whether it is always the case that we can assume a Gaussian action space and show that in the case of rotation (and subsequently 6D pose) it indeed beneficial to deviate from the standard Gaussian parameterization. To that end, we examine the two most common continuous control algorithms: SAC~\cite{haarnoja2018soft} and PPO~\cite{schulman2017proximal}, and replace the prediction of the mean and standard-deviation of a Gaussian, with the prediction of the parameters of a Bingham Distrbution (Figure \ref{fig:front_summary}), which is a distribution well suited for quaternions, and has had success in the supervised learning literature~\cite{gilitschenski2019deep, peretroukhin2020smooth}. When evaluating our approach on a Wahba problem environment and a set of vision-based robot manipulation tasks from RLBench~\cite{james2019rlbench}, we achieve superior performance over a Gaussian parameterization. Our hope is that the paper can encourage more research into developing other policy parameterization that are more suited for particular environments, rather than always assuming Gaussian.

\section{Related Work}

Continuous control RL policies are commonly parameterized as Gaussians with diagonal covariance matrices~\cite{schulman2017proximal,schulman2015trust,haarnoja2018soft}, though other parameterizations have been considered, including Gaussians with covariance matrix via the Cholesky factor~\cite{abdolmaleki2018maximum}, Gaussian mixtures~\cite{wulfmeier2019compositional}, Beta distributions~\cite{chou2017improving}, and Bernoulli distributions~\cite{seyde2021bang}. Rather than directly outputting continuous values, an alternative way of parameterizing a continuous control policy is via discretization, whether through growing action spaces~\cite{farquhar2020growing} or coarse-to-fine networks~\cite{james2021coarse}. All of these works share a common goal of moving away from the conventional Gaussian parameterization, but none are ideal when faced with an action space that requires rotation predictions. 

The decision to use the Bingham distribution for rotation and pose estimation is well studied for both orientation~\cite{gilitschenski2015unscented} and pose~\cite{glover2012monte, glover2014tracking, srivatsan2016estimating}. Inspired by these works, this paper extends the use of the Bingham distribution to the RL community; though in more general terms, we also use it as a way to stress that easy performance gains could be gained simply by correctly parameterizing the continuous control RL policies with the right distribution, rather than always assuming Gaussian. An alternative to using the Bingham distribution for rotation and pose estimation, is the $5D$ or $6D$ rotation parametrization presented in \citet{zhou2019continuity}, and later extended in \citet{labbe2020cosypose}; however, it is unclear how to represent this as a distribution --- a requirement for stochastic reinforcement learning.

The motivation of our work comes strongly from both \citet{peretroukhin2020smooth} and \citet{gilitschenski2019deep}, who show that the Bingham distribution can be used to model rotation uncertainty in a supervised learning setting. Due to the supervised learning domain, only parts of the distribution needed to be modeled, however in a reinforcement learning domain, we are required to fully estimate the Bingham distrubution and allow sampling, log probability of actions, and entropy calculation. The result is an RL algorithm that can fully model the Bingham distribution and allow for more natural exploration of 3D rotations.

\section{Background}

\begin{figure}
     \centering
     \includegraphics[width=\linewidth]{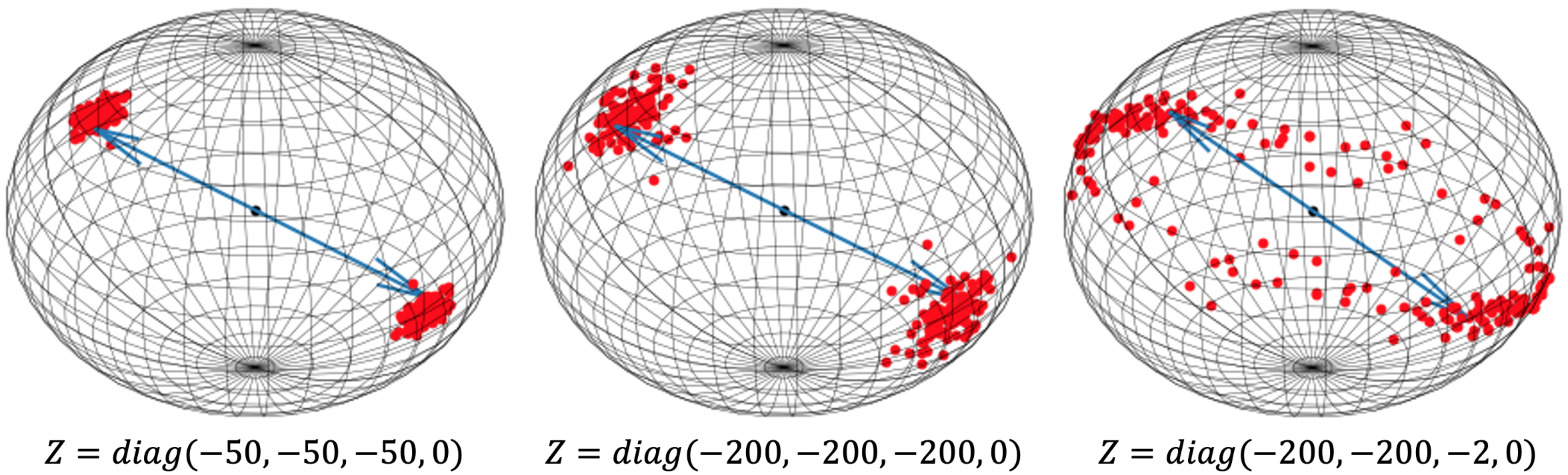}
     \caption{Illustrative example of how the Bingham distribution changes as the dispersion coefficients (given by $\diag{z_1, z_2, z_3, 0}$ with $z_1 \leq z_2 \leq z_3 \leq 0$) alter for a fixed $\Matrix{M}$.}
     \label{fig:zparams_example}
\end{figure}

For the proposed method, understanding of two key concepts are needed: (1) the Bingham distribution and the parameters that will need to be estimated, and (2) rejection sampling, as this is how we will be sampling from our policy during exploration. In this section, we outline these concepts, along with a brief summary of the reinforcement learning paradigm.

\subsection{Bingham Distribution}
\label{sec:bingham_distribution}

Rotation matrices, Axis-angles, and Quaternions are widely used representations for 3D rotation; these representations differ from other rotation representations, such as Euler angles, in that they do not suffer from Gimbal lock. Quaternions are often preferred over rotation matrices due to their compact nature: represented via a vector $q \in \Real^4$. Quaternions are naturally assumed to be have unit norm: i.e. $||q|| = 1$, and are antipodal symmetric, such that $q$ and $-q$ represent the same rotation. 

The Bingham distribution~\cite{bingham1974antipodally} naturally captures the properties of quaternions, and is derived from a zero-mean Gaussian conditioned to lie on the unit hypersphere, $S^d$. The probability density function is defined by:

\begin{align}
 p(\Vector{x}; \Matrix{M}, \Matrix{Z}) &= \frac{1}{N(\Matrix{Z})} \exp{\left( \Vector{x}^\T \Matrix{M} \Matrix{Z} \Matrix{M}^\T \Vector{x}\right)},
\end{align}
where $\Vector{x} \in S^3$, $N(\Matrix{Z})$ is a normalization constant, $\Matrix{M} \in \Real^{4 \times 4}$ is an orthogonal matrix, and $\Matrix{Z} \in \Real^{4 \times 4}$ is a matrix of dispersion coefficients, given by $\diag{z_1, z_2, z_3, 0}$ with $z_1 \leq z_2 \leq z_3 \leq 0$; these restrictions are in place for numerical and representational convenience. Intuitively, $\Matrix{M}$ holds information regarding the direction of the probability mass, while $\Matrix{Z}$ controls the spread along the direction, where smaller magnitudes imply larger spreads, and vice-versa; a visualization of this effect can be seen in Figure \ref{fig:zparams_example}. An important property of the Bingham distribution is that $\BinghamDistribution({\Matrix{M}}, {\Matrix{Z}}) = \BinghamDistribution({\Matrix{M}}, {\Matrix{Z} + c \Identity})$ for all $c \in \Real$, where $\Identity \in \Real^{4\times4}$ denotes identity matrix.

The normalization constant $N(\Matrix{Z})$ is a hypergeometric function that is non-trivial to compute, and defined as:

\begin{align}
 N(\Matrix{M} \Matrix{Z} \Matrix{M}^\T) &= \int_{||x|| = 1} \exp{\left( \Vector{x}^\T \Matrix{M} \Matrix{Z} \Matrix{M}^\T \Vector{x}\right)} d\Vector{x} \\
 N(\Matrix{Z})  &= \int_{||x|| = 1} \exp{\left( \Vector{x}^\T \Matrix{Z} \Vector{x}\right)} d\Vector{x}.
\label{eq:normconst}
\end{align}
Note that due to the transformation theorem and the orthogonality of M: $N(\Matrix{M} \Matrix{Z} \Matrix{M}^\T) = N(\Matrix{Z})$. Computation of the normalization constant is non trivial and becomes more difficult as the dispersion $\Matrix{Z}$ decreases; it is an active area of research in its own right, with some solutions including series expansions~\cite{koev2006efficient}, saddle point approximations~\cite{kume2005saddlepoint}, holonomic gradient descent~\cite{koyama2014holonomic}, and precomputed lookup tables~\cite{glover2013libbingham}. In section \ref{sec:probability_density_function} we describe how we calculate the constant.

\subsection{Rejection Sampling}

Rejection sampling is a technique used to generate observations from a distribution. Assume two densities:

\begin{align}
f(x) = c_f f^{*}(x), \ g(x) = c_g g^{*}(x),
\end{align}
where $f^{*}$ and $g^{*}$ are known functions. Suppose it is possible to simulate easily from $g$ (envelope function) and it is desired to simulate observations from $f$ (target function). The key requirement is that there is a known bound of the form:

\begin{align}
f^{*}(x) \leq C^{*}g^{*}(x) \ \forall x,
\end{align}
for some constant $C^{*}$. The rejection sampling algorithm proceeds as follows:

\begin{enumerate}
    \item Simulate $X \sim g$ independently of $W \sim \text{Unif}(0, 1)$.
    \item If $W > \frac{f^{*}(X)}{C^{*}g^{*}(X)}$, then reject $X$ and return to step 1, else $X$ is a sample from the desired distribution.
\end{enumerate}

\subsection{Reinforcement Learning}

The reinforcement learning paradigm assumes an agent interacting with an environment consisting of states $\bs \in \states$, actions $\ba \in \actions$, and a reward function $R(\st,\at)$, where $\st$ and $\at$ are the state and action at time step $t$ respectively. The goal of the agent is then to discover a policy $\pi$ that results in maximizing the expectation of the sum of discounted rewards: $\E_\pi [\sum_t \gamma^t R(\st, \at)]$, where future rewards are weighted with respect to the discount factor $\gamma \in [0, 1)$. In this paper, we apply our Bingham Policy Parameterization (BPP) to three different algorithms: Soft-actor critic (SAC)~\cite{haarnoja2018soft}, proximal policy optimization (PPO)~\cite{schulman2017proximal}, and attention-driven robotic manipulation (ARM)~\cite{james2021attention}. We briefly outline these below.

\textbf{SAC} is a maximum entropy off-policy RL algorithm that, in addition to maximizing the sum of rewards, also maximizes the entropy of a policy: 
\begin{align}
\E_\pi [ \sum_t \gamma^t [ R(\st, \at)+\alpha \mathcal{H} (\pi (\cdot | \st ) ) ] ],
\end{align}
where $\alpha$ is a temperature parameter that determines the relative importance between the entropy and reward.

\textbf{PPO} is a policy gradient method that imposes policy ratio $r(\theta) = \pi_{\theta}(\st, \at)/\pi_{\theta_{\text{old}}}(\st, \at)$ to stay within a small interval $\epsilon$ around 1: 
\begin{align}
\E_\pi [ \min ( r(\theta) A_{\theta_{\text{old}}}, \text{clip} (r(\theta), 1 - \epsilon, 1 + \epsilon) A_{\theta_{\text{old}}}) ],
\end{align}
where $A$ is an advantage function.

\textbf{ARM} introduced several core concepts that facilitate the learning of robot manipulation tasks. These included Q-attention, keypoint detection, demo augmentation, and a high-level next-best pose action space. Given an observation, $\obs$ (consisting of an RGB image, $\rgb$, an organised point cloud, $\pcd$, and proprioceptive data, $\proprio$), the Q-attention module, $\qattn_\qattnp$, outputs 2D pixel locations of the next area of interest. This is done by extracting the coordinates of pixels with the highest value: $(x, y) = \argmaxtwod_{\ba'} \qattn_{\qattnp}(\obs, \ba')$, where $\argmaxtwod$ is an \textit{argmax} taken across two dimensions. These pixel locations are used to crop the RGB image and organised point cloud inputs and thus drastically reduce the input size to the next stage of the pipeline; this next stage is an actor-critic next-best pose agent using SAC as the underlying algorithm. For further details on keypoint detection and demo augmentation, we point the reader to~\cite{james2021attention}.

\section{Bingham Policy Parameterization}
\label{sec:bpp}

We now present our Bingham Policy Parameterization (BPP), which can be applied to any stochastic RL policy. The Bingham distribution parameters $\Matrix{M}$ and $\Matrix{Z}$ are learned end-to-end directly from raw observation data. Similarly to~\cite{gilitschenski2019deep}, we predict a 19 dimensional vector $\Vector{v} \in \Real^{19}$, which we then convert to $\Matrix{M}$ and $\Matrix{Z}$ through differentiable transforms $\mathcal{T}_{\Matrix{M}}$ and $\mathcal{T}_{\Matrix{Z}}$. Transform $\mathcal{T}_{\Matrix{Z}}$ is defined as $\mathcal{T}_{\Matrix{Z}}(v_1, v_2, v_3) = \diag{z_3, z_2, z_1, 0}$, where $z_i = -\sum_{k=1}^{i} \exp{v_k}$; this satisfies the Bingham property that $z_1 \leq z_2 \leq z_3 \leq 0$. 

For transform $\mathcal{T}_{\Matrix{M}}$, we subdivide $v_4, \dots, v_{19}$ into four vectors $\Vector{v}^i \in \Real^4$, for $i = 1, \dots, 4$. The Gram-Schmidt orthonormalization method is then applied to these vectors to obtain $\mathcal{T}_{\Matrix{M}}(v_4, \dots, v_{19}) = [\hat{\Vector{m}}_1, \hat{\Vector{m}}_2, \hat{\Vector{m}}_3, \hat{\Vector{m}}_4]$, where $\hat{\Vector{m}}_i = \text{Norm}(\Vector{v}_i - \sum_{k=1}^{i-1}  \langle \hat{\Vector{m}}_k, \Vector{v}_i\rangle \cdot \hat{\Vector{m}}_k)$, for $i = 1, \dots, 4$, and where $\text{Norm}$ signifies vector normalization. During our initial experimentation, we evaluated training stability of two orthonormalization: classical Gram-Schmidt and modified Gram-Schmidt, and found no substantial difference. The experiments in this paper use classical Gram-Schmidt.

\subsection{Probability Density Function}
\label{sec:probability_density_function}

The final piece needed for the Bingham p.d.f is the normalization constant $N(\Matrix{Z})$. However, $N(\Matrix{Z})$ is a hypergeometric function that is non-trivial to compute. Earlier work for rotation uncertainty approximated this with a lookup table based interpolation mechanism~\cite{gilitschenski2019deep}, however, due to the fact that RL algorithms often require many samples to optimize (often in the order of millions), we found that the finite-difference gradient calculation for the lookup table based interpolation was too slow. Therefore, we instead approximate $N(\Matrix{Z}) \approx f_N(\Matrix{Z})$ and $\Delta_z N(\Matrix{Z}) \approx \Delta_z f_N(\Matrix{Z})$ via a lightweight fully-connected neural network, allowing for fast inference and gradient calculation. To achieve this, we build a pre-computed dataset for $N(\Matrix{Z})$ for many values of Z using Equation \ref{eq:normconst}, and then use this to train the neural network. 
As mentioned in Section \ref{sec:bingham_distribution}, many methods exist for solving the normalization constant. Given that we are computing the dataset offline, we opt for the slower, but more accurate numerical integration solution. This follows \textit{Gilitschenski et al.}~\cite{gilitschenski2019deep} where Scipy’s tplquad method is used to compute a triple integral for a range of Z values interpolated from $-500$ to $0$. Before considering the full integral, first note that we need to transform the unit quaternions to 4D spherical coordinate via:

\begin{align}
t(\phi_1, \phi_2, \phi_3) = 
\begin{bmatrix}
\sin(\phi_1) \sin(\phi_2) \sin(\phi_3) \\
\sin(\phi_1) \sin(\phi_2) \cos(\phi_3) \\
\sin(\phi_1) \cos(\phi_2) \\
\cos(\phi_1)
\end{bmatrix}
\end{align}

and corresponding volume correction term $dx = \sin(\phi_2) \sin^{2}(\phi_3) d\phi_1 d\phi_2 d\phi_3$, where $\phi_1, \phi_2 \in [0, \pi]$ and $\phi_3 \in [0, 2\pi]$. With this, the actual integral can be computed as:

\begin{align}
 N(\Matrix{Z}) &= \int_{||x|| = 1} \exp{\left( \Vector{x}^\T \Matrix{Z} \Vector{x}\right)} d\Vector{x} \\
               &= 
               \begin{aligned}
               \int_{0}^{2\pi} \int_{0}^{\pi} \int_{0}^{\pi} & \exp{\left( t(\phi_1, \phi_2, \phi_3)^\T \Matrix{Z} t(\phi_1, \phi_2, \phi_3)\right)} \\ 
               & \sin(\phi_2) \sin^{2}(\phi_3) d\phi_1 d\phi_2 d\phi_3
               \end{aligned}
\end{align}

Note that the pre-training phase for the normalization constant is not only very quick (in the order of minutes to train), but also is a fixed cost burdened to us, the authors. The same trained network can run in any environment, and it will never see out of distribution data, as the Z value input will always lie within $-500$ to $0$. Therefore, users of this policy parameterization do not need to perform any pre-training when using our supplied weights, and so the method can be seen as having no pre-training requirement for the end user.

\subsection{Sampling}

Crucial to a stochastic policy is the ability to sample in a differentiable way. For this, we use an acceptance-rejection method using the angular central Gaussian distribution (distribution of a normalized Gaussian random variable) as an envelope~\cite{kent2013new}. First, recall the angular central Gaussian distribution, $ACG(\Omega)$:

\begin{align}
f_{ACG}(x) = \sqrt{|\Omega|} (x^\T \Omega x)^{-\frac{q}{2}}
\end{align}

\begin{figure}
     \centering
     \includegraphics[width=\linewidth]{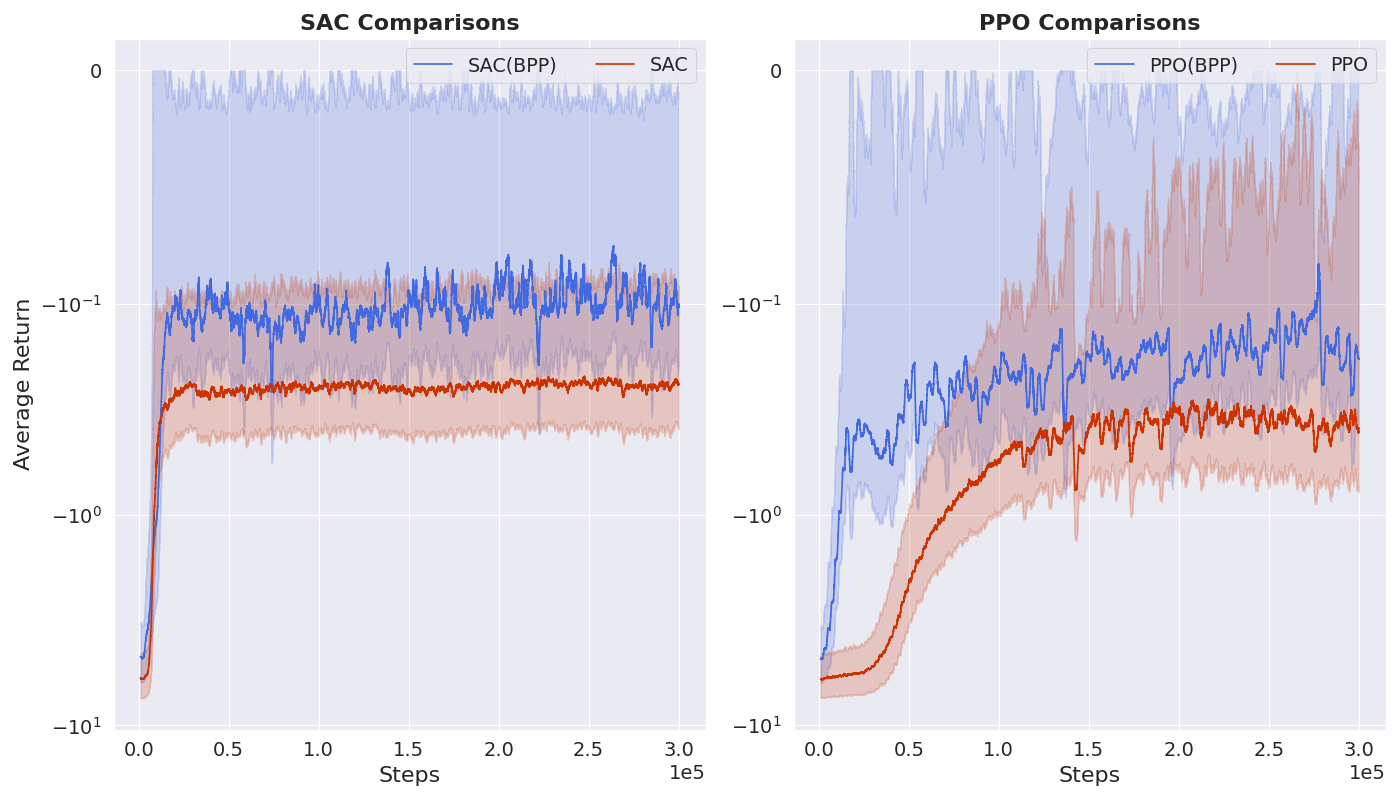}
     \caption{Wahba environment results PPO and SAC, with and without the BPP. Solid lines represent the average evaluation over 5 seeds, where the shaded regions represent the $std$.}
     \label{fig:wahba_results}
\end{figure}

The distribution is named by the fact that if $y \sim \NormalDistribution{0}{\Sigma}$ where $\Sigma$ is positive definite, then $\frac{y}{||y||} \sim ACG(\Omega)$. The goal here is to choose an appropriate $\Omega$ that depends on $Z$. We use the general inequality given in \textit{Kent et al.}~\cite{kent2013new} that uses the concavity of the log function to help in the construction of rejection sampling algorithms. The inequality is given as a function of $u$:

\begin{align}
\exp(-u) &\leq \exp(-(n-b)/2) \Big( \frac{n/b}{1 + 2u/b} \Big)^{n/2},
\label{eq:env_ineq}
\end{align}
where $n > 0$ is the number of dimensions in euclidean space ($n=4$ for quaternions) and $b$ is a fixed constant: $0 < b < n$. Setting $u = x^{\T}\Matrix{A}x$ in Equation \ref{eq:env_ineq} and setting $\Omega = \Omega(b) = \Identity + \frac{2\Matrix{A}}{b}$, where $\Matrix{A} = \Matrix{M}\Matrix{Z}\Matrix{M}^{\T}$ for brevity. This yields:

\begin{align}
\exp(x^\T \Matrix{A} x) &= \exp(-u) \\
                       &= \exp(-(n-b)/2) \Big( \frac{n/b}{1 + 2x^\T \Matrix{A} x/b} \Big)^{n/2}  \\
                       &= \exp(-(n-b)/2) \Big( \frac{n/b}{x^\T \Omega x} \Big)^{n/2}.
\end{align}

The corresponding bound $C(b)$ takes the form:

\begin{align}
C(b) &= \frac{1}{N({\Matrix{Z}})} \exp(-(n-b)/2) (n/b)^{n/2} |\Omega(b)|^{-1/2},
\end{align}
where the aim is to find the optimal bound $C(b_0)$ via the solution of: 

\begin{align}
\sum_{i}^{n} \frac{1}{b + 2z_i} = 1.
\label{eq:solve_b}
\end{align}

This can be done via Scipy's \textit{fsolve} function. Note, however, that the calculation of b is non-differentiable; in the pursuit of fast sampling and gradient calculation, we opt for a similar method as was taken for calculating the normalization constant; that is, we approximate the solution to Equation \ref{eq:solve_b} with $f_b(Z)$ via a small fully-connected neural network. Similarly to Section \ref{sec:probability_density_function}, we build a pre-computed dataset for $b(\Matrix{Z})$ for many values of Z by numerically solving Equation \ref{eq:solve_b}, and then use this to train the neural network.

Because the rejection sampling relies on samples from the angular central Gaussian (ACG) distribution, and because differentiable sampling from a multivariate Gaussian is well understood (via the reparameterization trick), we can also achieve differentiable sampling from the Bingham distribution. The rejection sampling outlined above was found to have a high acceptance rate of $>90\%$ throughout training. The implementation in PyTorch consists of sampling $B$ number of samples from the (ACG) distribution, sorting them according to the rejection criteria, and choosing the first element. In practice, with $B=10$, there is always an accepted sample in each batch. Given the sample efficiency, more sophisticated sampling, such as MCMC was not considered, however this would make for interesting future work. Note that \textit{Kent et al.}~\cite{kent2013new} found that the rejection sampling method outlined above superseded an MCMC method~\cite{kume2006sampling} for the Bingham distribution.

\subsection{Entropy}

Another crucial component for parametrizing the policy with a Bingham distribution is the ability to compute analytic entropy for exploration. The entropy of the Bingham distribution $\mathcal{B}$ is:

\begin{align}
 \mathcal{H}(\mathcal{B}) &= -\int_{\Vector{x} \in \Real^4} \mathcal{B}(\Vector{x}; \Matrix{M}, \Matrix{Z}) \log(\mathcal{B}(\Vector{x}; \Matrix{M}, \Matrix{Z})) \\
 &= \log(N(\Matrix{Z})) - Z \frac{\Delta N(\Matrix{Z})}{N(\Matrix{Z})}.
\end{align}

Note that we are again presented with the need to calculate the normalization constant and its gradient; for this, we utilize our fully-connected neural network approximation $f_N(\Matrix{Z})$. Note that an analytic entropy calculation is not always needed, and instead can be estimated via the negative log probability of actions output by the policy; however, calculating entropy explicitly does often lead to improved training stability.

\section{Experimental Results}

Our experiments aim to answer the following questions: \textbf{(1)} Does swapping the standard Gaussian policy parameterization (GPP) with a Bingham policy parameterization (BPP) improve performance when the action space only involves 3D rotation? \textbf{(2)} How does the our method perform when the action space becomes 6D poses, where the rotation part uses a BPP and the translation part uses a GPP? \textbf{(3)} Does the BPP also extend to more practical, sparsely rewarded vision-based manipulation tasks? We answer these questions through a series of simulated experiments.

\subsection{Rotation: Wahba Problem}

\begin{figure}
     \centering
     \includegraphics[width=0.48\textwidth]{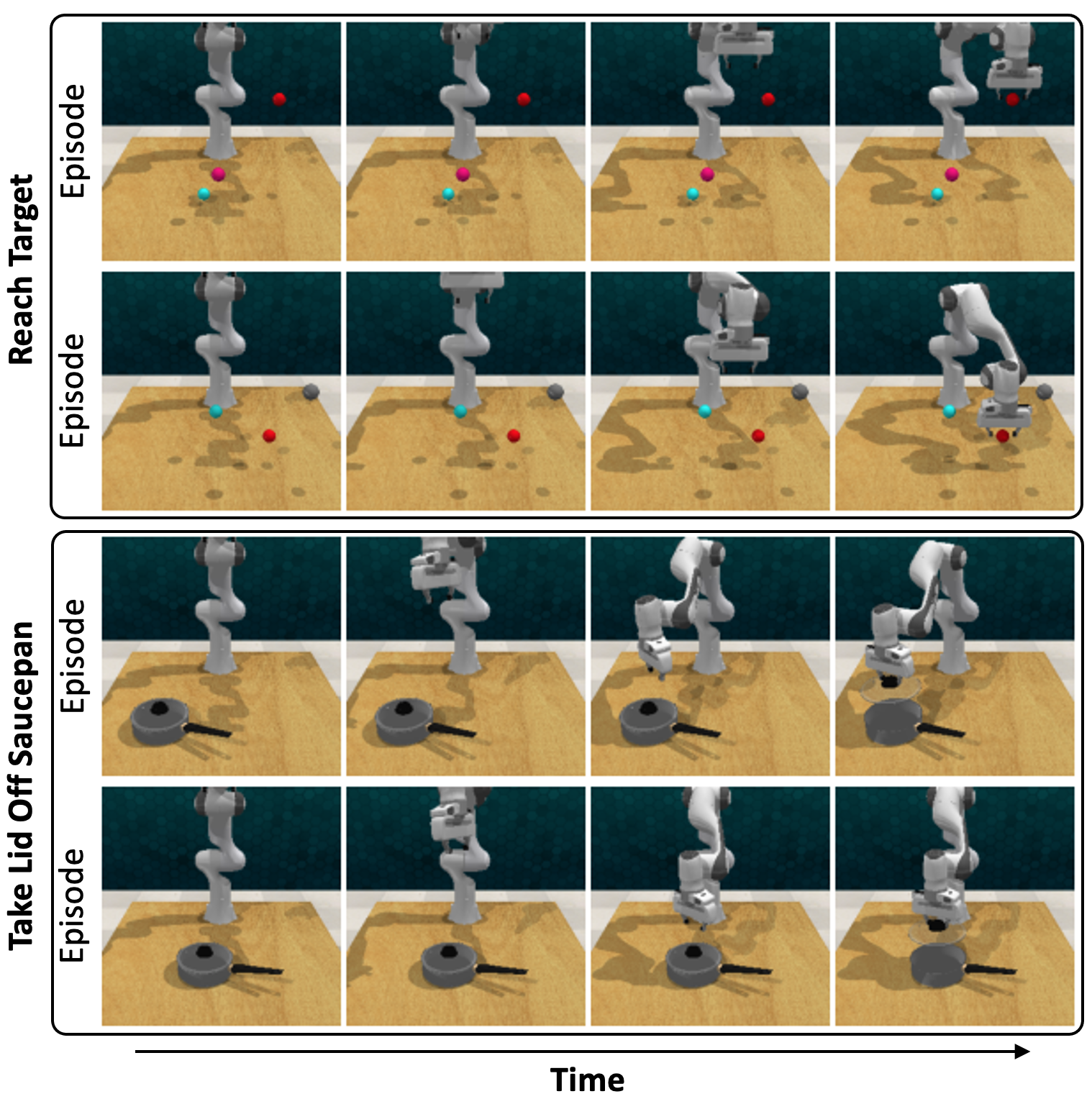}
     \caption{The two RLBench tasks used in Section \ref{sec:lowdim}. The reach target task requires the end-effector to reach for the red target, while the take lid off saucepan task requires the saucepan lid to be grasped and lifted above the saucepan. Note that the positions of objects are placed randomly at the beginning of each episode.}
     \label{fig:rlbench_tasks}
\end{figure}

The first set of experiments aim to asses the performance gain when moving from a GPP to a BPP for an environment that requires only rotation actions. To keep the experiment as simple a possible, we look to the Wahba problem~\cite{wahba1965least}. Proposed in 1965, the Wahba problem seeks to find a rotation matrix between two coordinate systems from a set of (noisy) weighted vector observations. It has previously been used as the initial experiment when comparing rotation prediction methods~\cite{zhou2019continuity,peretroukhin2020smooth}, and so here we adapt the Wahba problem for the reinforcement learning domain. The environment is modeled as a one-step MDP, where the aim is to produce a rotation $\Rotation$ such that:

\begin{equation}
\Vector{v}_i = \Rotation \Vector{u}_i + \Vector{\epsilon}_i~, ~~ \Vector{\epsilon}_i \sim \NormalDistribution{\Vector{0}}{\sigma^2 \Identity}.
\end{equation}

The vectors $\Vector{u}_i$ and $\Vector{v}_i$ are given as observations, the reward is defined as a negated chordal loss~\cite{hartley2013rotation}, and the expected action is a unit quaternion. Note that although this task is inspired from the supervised learning domain, we do not claim or aim to solve Wahba's problem better than supervised learning, but rather to compare BPP and GPP, and show that BPP is beneficial to use when an environment requires a rotation action space. Both BPP and GPP use the same simplified PointNet~\cite{qi2017pointnet} style architecture used in~\cite{zhou2019continuity,peretroukhin2020smooth}, along with the default hyperparameters in Stable Baselines3~\cite{stablebaselines3}.

We present the results of the Wahba reinforcement learning (RL) environment in Figure \ref{fig:wahba_results}. Results show that for both SAC~\cite{haarnoja2018soft} and PPO~\cite{schulman2017proximal} perform best when using a BPP, converging faster as well as gaining a better final performance. We conclude that for rotation, it seems that by simply correctly modeling the action space distribution, easy performance gains can be made without having to make any changes to the underlying RL algorithm. Code for these experiments have been included in the supplementary material; Stable Baselines3~\cite{stablebaselines3} was used to make this experimental setup as reproducible as possible. 

\subsection{Pose: Low-dimensional Robotic Manipulation}
\label{sec:lowdim}

\begin{figure}
     \centering
     \includegraphics[width=\linewidth]{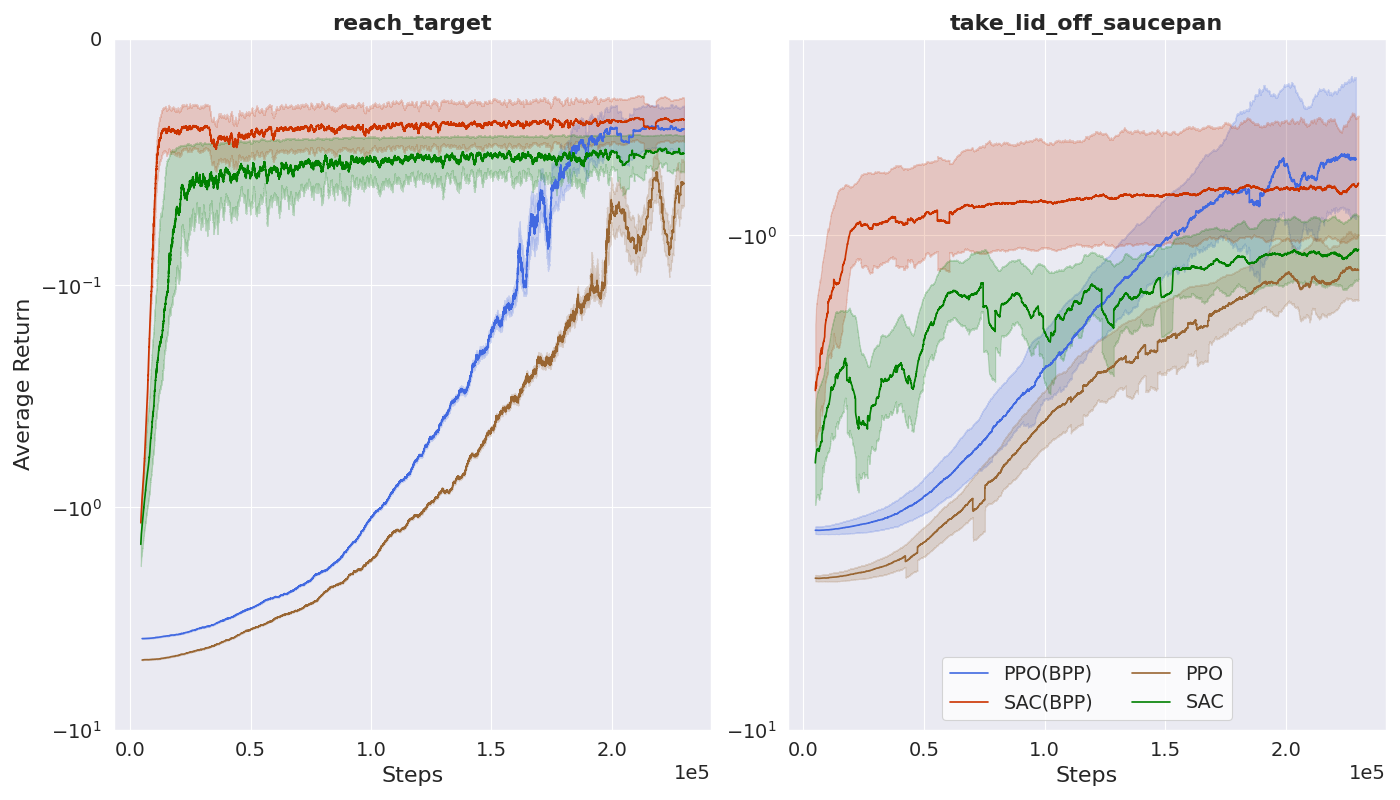}
     \caption{RLBench environment (full state and shaped rewards) results for PPO and SAC, with and without the BPP. Here, two tasks are considered: \textit{reach\_target} and \textit{take\_lid\_off\_saucepan}. Solid lines represent the average evaluation over 5 seeds, where the shaded regions represent the $std$.}
     \label{fig:full_state_rlbench}
\end{figure}

The Wahba experiments showed that BPP is preferred when using action spaces that require rotation. However, rotations are often predicted in tandem with translation to give 6D poses. Therefore, in this set of experiments, we aim to investigate an RL environment which requires a pose action space. A natural fit for this would be a robot manipulation environment, where the agent is tasked with outputting a series of next-best poses to complete a task. For these tasks, we look to RLBench~\cite{james2019rlbench}. RLBench is a large-scale benchmark and learning environment designed to facilitate research in vision-guided manipulation research. RLBench comes with a number of action modes; for our experiments we use `\textit{EndEffectorPoseViaPlanning}', where a linear path (via IK) or non-linear path (via sample-based planning) is found to bring the robot end-effector to the given target next-best pose. 

\begin{figure*}
     \centering
     \includegraphics[width=\linewidth]{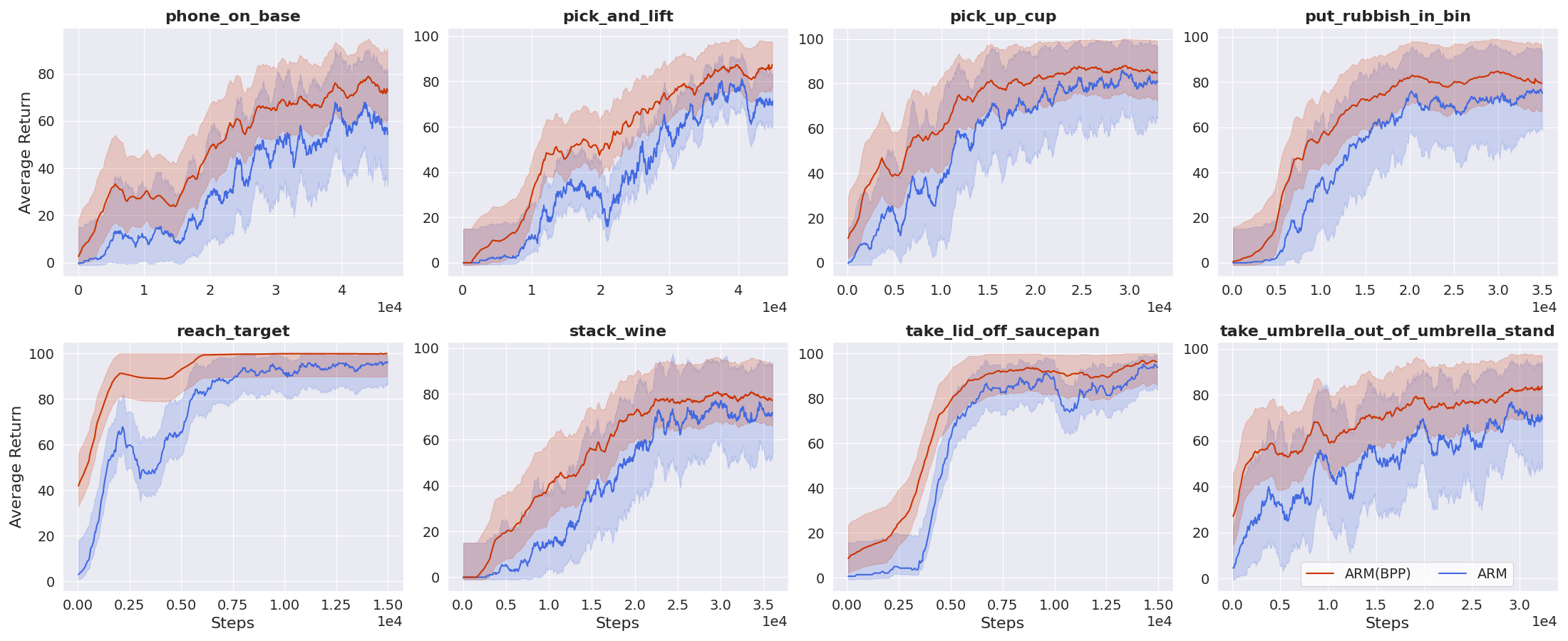}
     \caption{RLBench environment with high-dimensional state (RGB \& point cloud) and sparse reward on task completion. All methods receive 100 demos which are stored in the replay buffer prior to training. Solid lines represent the average evaluation over 5 seeds, where the shaded regions represent $std$.}
     \label{fig:arm_bpp}
\end{figure*}

For these initial experiments, we will remove some of the challenging part of RLBench, including high-dimensional vision observation (e.g. RGB, Depth. etc) and sparse rewards, and instead supply full-state information as well as shaped rewards. Note that in the next section, all of these challenging aspect are re-added. We chose to experiment on two tasks: \textit{reach\_target} and \textit{take\_lid\_off\_saucepan}, which we briefly describe below:

\noindent \textbf{\textit{reach\_target}} requires the gripper to move to the red target. The observation is the 3D translation of the target sphere along with the current 6D pose of the arm. The reward function is the negative L2 distance between the target sphere and the current end-effector position. 

\noindent \textbf{\textit{take\_lid\_off\_saucepan}} is more challenging, involving the use of the gripper and a longer task horizon, where the agent must learn to grasp the saucepan handle and lift it above the pan. The observation is the pose of the saucepan handle along with the current 6D pose of the end-effector. The reward function is the negative L2 distance between the saucepan handle and the arms current end-effector position plus the negative distance of the saucepan to the target height. 

Two example rollouts of these tasks can be found in Figure \ref{fig:rlbench_tasks}. All methods use the same default network architecture and hyperparameters within Stable Baselines3~\cite{stablebaselines3}.

A notable part of these experiments is the use of both the Bingham and Gaussian policy parameterizations, where SAC(BPP) and PPO(BPP) use the BPP for rotation, and the GPP for translation and gripper; this is done by predicting an output of size $27$, where $19$ values are used for parameterizing the Bingham (as laid out in Section \ref{sec:bpp}), $6$ values are used to parameterize the translation Gaussian ($3$ each for mean and log std), and $2$ values are used to parameterize the gripper Gaussian ($1$ each for mean and log std). Standard SAC and PPO (without BPP) predicts $14$, where $8$ values are used to parameterize the rotation Gaussian ($4$ each for mean and log std), $6$ values are used to parameterize the translation Gaussian, and $2$ values are used to parameterize the gripper Gaussian. Note that the gripper could be represented via a categorical distribution, however this is out of scope for this paper. Results for this section are shown in Figure \ref{fig:full_state_rlbench}, where the gains seen for rotation only in the Wahba problem are also clearly present in the next-best pose environment.

\subsection{Pose: Image-based Robotic Manipulation}
\label{sec:highdim}

Finally, we move beyond the two toy-problems presented in the previous two sections, and show how our method scales in a number of ways: \textbf{(1)} to vision-based tasks, \textbf{(2)} to sparse rewarded tasks, and \textbf{(3)} to a more sophisticated robot learning algorithm. For these experiments, we use the same 8 RLBench tasks as used in \citet{james2021attention}. Note, unlike the previous section, we do not modify them to have shaped rewards or low-dimensional observations (i.e. RLBench in its challenging default state); making the environment significantly harder. Given that PPO and SAC are unable to perform well in this domain (due to increased challenge of sparse rewards and image observations), we instead make use of ARM~\cite{james2021attention}. ARM is a state-of-the-art manipulation algorithm that introduced several core concepts which facilitate the learning of robot manipulation tasks given only a small number of demonstrations; these include Q-attention, keyframe discovery, and demo augmentation. SAC is used as the underlying RL algorithm within ARM, and so the only change we make is to swap the policy parameterization to use BPP rather than GPP, just as was done in the previous section. As was done in ARM~\cite{james2021attention}, we fill the buffer with 100 demonstrations and apply keyframe discovery and demo augmentation. The exact same architecture and hyperparameters are used as described in~\cite{james2021attention}. The results of ARM compared to ARM(BPP) are shown in Figure \ref{fig:arm_bpp}. The results show that the gains seen in the easier experimental setup (with shaped-reward and full state) scale to the much more challenging setup (with sparse rewards and high-resolution observations). Note that it is not often that new RL algorithms also run experiments on environments with sparse reward and high-dimensional observations; these are usually left for future work, leaving the reader unsure of its scaleability to the image domain.

\section{Conclusion}

In this paper we have explored an alternative policy parameterization for representing 3D rotations during reinforcement learning via the Bingham distribution. Our results have shown that by using a Bingham distribution over the commonly used Gaussian distribution yields superior results in a range of environments that require either rotation action spaces or pose action spaces. Indeed, we hypothesis that this could spur research in other RL sub-domains to experiment with other policy parameterizations that could yield similar gains by considering alternatives to the common Gaussian parameterization. 

Despite the positive results, there are undoubtedly weaknesses. Although the normalization constant and $b$ constant approximation via a neural network is simple, it is likely to introduce errors when sampling and acquiring the log probability of actions, which may cause underlying stability issues for the RL process; the Gaussian parameterization does not suffer from this weakness. Avenues for future work include alternative methods for estimating the normalization constant and $b$ constant approximation. We are also interested in looking more closely at the pose action space in particular; investigating alternatives for representing the translation and rotation component jointly under one distribution, rather than representing rotation via the Bingham and translation via a Gaussian. 

\section*{Acknowledgments}

This work was supported by the Hong Kong Centre for Logistics Robotics.

\bibliographystyle{plainnat}
\bibliography{ref}

\end{document}